\documentclass[letterpaper, 10 pt, conference]{ieeeconf}  

\IEEEoverridecommandlockouts                              

\usepackage{graphicx} 
\usepackage{epsfig} 
\usepackage{url}
\usepackage{hyperref}

\title{\LARGE \bf
ROXIE: Defining a Robotic eXplanation and Interpretability Engine
}

\author{Francisco J. Rodríguez-Lera$^{*,1}$ and Miguel A. González-Santamarta$^{1}$ and Alejandro González-Cantón$^{1}$ and\\ Laura Fernández-Becerra$^{1}$ and David Sobrín-Hidalgo$^{1}$ and Angel Manuel Guerrero-Higueras$^{1}$
\thanks{$^{*}${Correspondence to: \tt\small fjrodl@unileon.es}}%
\thanks{$^{1}$Robotics Group, University of León, Spain}%
}

\begin{document}

\maketitle
\thispagestyle{empty}
\pagestyle{empty}

\begin{abstract}
In an era where autonomous robots increasingly inhabit public spaces, the imperative for transparency and interpretability in their decision-making processes becomes paramount. This paper presents the overview of a Robotic eXplanation and Interpretability Engine (ROXIE), which addresses this critical need, aiming to demystify the opaque nature of complex robotic behaviors. This paper elucidates the key features and requirements needed for providing information and explanations about  robot decision-making processes. It also overviews the suite of software components and libraries available for deployment with ROS 2, empowering users to provide comprehensive explanations and interpretations of robot processes and behaviors, thereby fostering trust and collaboration in human-robot interactions.
\end{abstract}

\section{INTRODUCTION}



Nowadays, the quest for transparency and interpretability stands as a cornerstone for fostering trust and collaboration between humans and machines. As robots increasingly permeate public spaces, from airports to hospitals, it becomes imperative to demystify their decision-making processes. Robot explainability and interpretability offer pathways to achieving this transparency, enabling users, operators, and stakeholders to comprehend the rationale behind a robot's actions.

The significance of transparency in autonomous systems cannot be overstated. It not only engenders trust but also facilitates smoother integration of robots into human-centric environments. When individuals understand why and how a robot makes decisions, they are more likely to accept and collaborate with it. This symbiotic relationship between humans and robots underscores the necessity for developing robust mechanisms that elucidate the inner workings of autonomous systems.

In this paper, we embark on a technological exploration to identify and design a Robotic eXplanation and Interpretability Engine (ROXIE) tailored for autonomous robotic systems deployed in public spaces. Our objective is to delve into the fundamental aspects of explainability and interpretability in robotics, with a specific focus on their application in real-world scenarios. At this juncture in our research, the impact of ROXIE is conceptualized through its efficacy in providing transparency and interpretability pertaining to the decision-making processes of autonomous robots.

By delineating the rationale behind our endeavor, we aim to lay the groundwork for a comprehensive understanding of the role played by explainability and interpretability in autonomous robotics. Through the development of ROXIE, we aspire to bridge the gap between technical sophistication and human comprehension, thereby paving the way for a future where autonomous robots coexist with human counterparts in public spaces.

\subsection{Contribution}
This paper presents and illustrates the design of an Explicability and Interpretability framework to use in assessing the theoretical and practical contributions. The community should answer the next  

\begin{itemize}
    \item Expected requirements for an explainable and interpretable in any autonomous robot. 
    \item Identification of techniques and tools involved currently available off-the-shelf in a ROS 2 scenario.
\end{itemize}


The rest of the paper is presented as follows. Section\ref{sec:ROXIE} presents the overall overview of ROXIE. Section\ref{sec:TOOLKIT} overviews our initial complete flow using some of the ROS 2 tools available off-the-shelf for fulfilling the requirements required to deploy a ROXIE in a robot. Finally, Section \ref{sec:CONCLUSIONS} presents the conclusions of this work.

\section{ROXIE}
\label{sec:ROXIE}
The acronym ROXIE, which stands for "Robotic eXplanation and Interpretability Engine" encapsulates a holistic strategy aimed at confronting the inherent challenges posed by the opacity inherent in intricate robotic decision-making processes. In essence, ROXIE embodies a comprehensive and systematic approach to unraveling the intricacies and enhancing the transparency of complex decision-making within autonomous robotic systems. Thus, ROXIE defines three main toolkits

\begin{itemize}
    \item REFT: Robotic Explanation Framework Toolkit
    \item RIFT: Robotic Interpretability Framework Toolkit
    \item RAFT: Robotic Accountability Framework Toolkit
\end{itemize}

The term "Framework Toolkit" here refers to a collection of tools, resources, and components bundled together to facilitate the development, implementation, and maintenance of software applications. The key components and features of a Framework Toolkit may include code libraries, documentation, templates and normalized proposals, Integrated Development Environment (IDE), command line tools, testing utilities, and debugging. These components should be associated with community support. 

At the same time, the community faces the use of Explanation and Interpretability indistinctly. Using the Machine Learning approaches, Rudin~\cite{rudin2019stop} defines the boundary between interpretable and explainable machine learning (ML). The author presents the interpretable ML that prioritizes the creation of models inherently understandable by design, whereas explainable ML endeavors to furnish retrospective explanations for pre-existing black box models. For authors, Explainable Robotics 
follows Pennington’s model of explanation-based decision-making~\cite {pennington1993reasoning}. The idea here is that an explanation is presented as a mental representation of a situation relevant to a decision-making system that emphasizes events, conditions and relationships.

\begin{figure*}[ht]
    \centering
    \vspace{2mm}
    \includegraphics[width=0.99\textwidth]{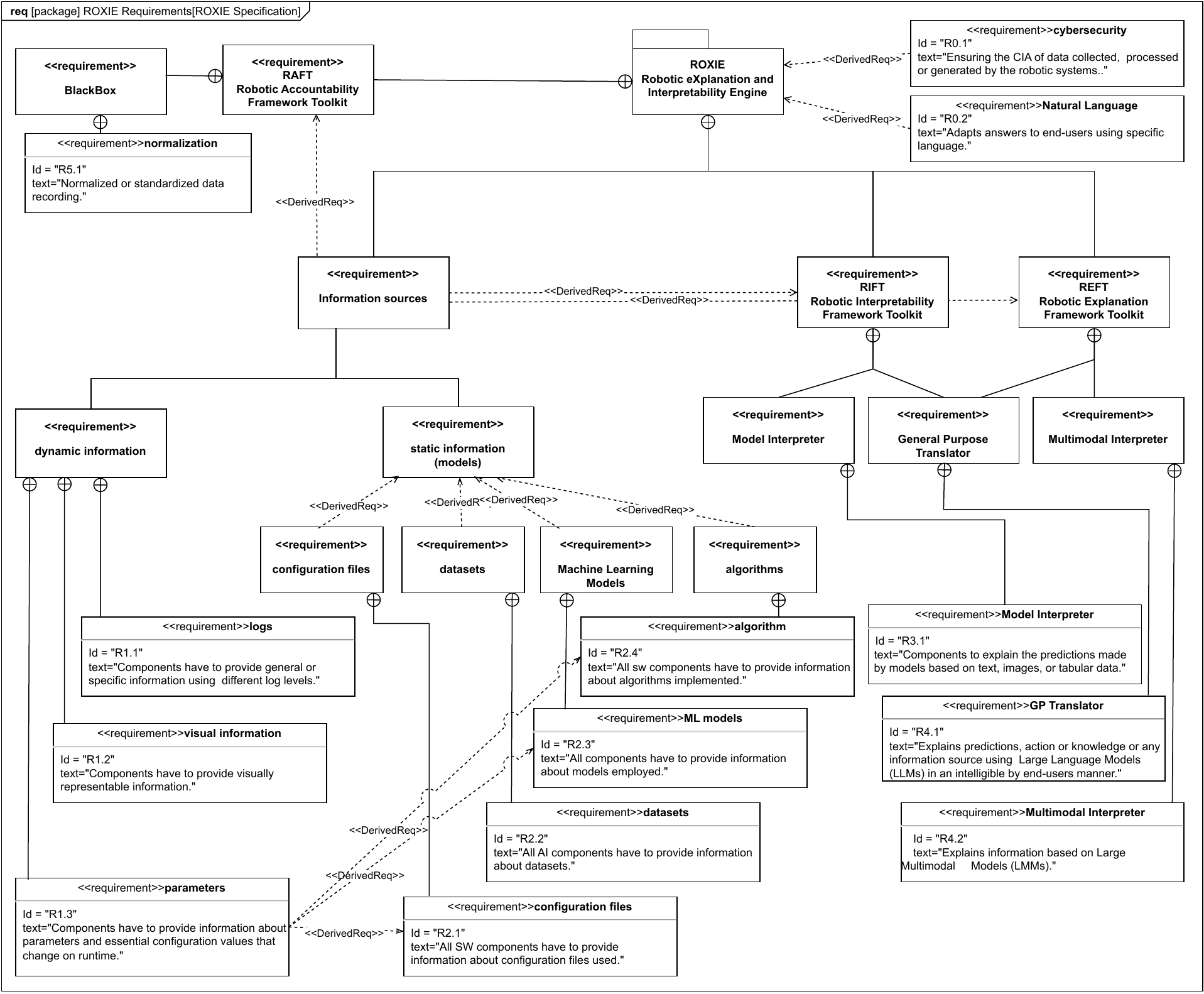}
    \caption{Robotic eXplanation and Interpretability Engine requirement definition.}
    \label{fig:ROXIE}
\end{figure*}

Figure~\ref{fig:ROXIE}  overviews the full set of requirements needed for fulfilling an explainable and Interpretable engine. The engine is proposed as a set of requirements instead of components. This is because some of them will be associated with many software components. 

There are two basic requirements that would be considered traversal to all others:

\begin{itemize}
    \item [R0.1] Cibersecurity: all ROXIE components, hardware and software, have to present clear basic principles of cibersecurity. They should have in mind confidentiality, integrity and availability of all data collected, processed or generated. This includes approaches such as antitampering, online backup, and physical protection among others. 

    \item [R0.2] Natural Language: ROXIE has to present information in a natural manner, trying to use natural language. Crypted logs, non-visualizable information or non-logged robot behaviors could convert the robot into a non-trustable platform.  

\end{itemize}

Roboticists and developers have to keep these two requirements in mind in order to be able to offer trustable robots. A modified robot is a non-trustable robot, and therefore, would be complicated to explain the reasons that generate a specific behavior or misguide end-users to its real performance. Besides, unclear messages or technical-specific information can confuse individuals interacting with robots.

\subsection{INFORMATION SOURCES}

This is the main requirement of a robot. In an explainable and interpretable engine, the importance of information sources lies in ensuring transparency, accountability, and understanding of the system's decision-making process. 

Knowing the sources of information helps users, stakeholders, and regulators understand how decisions are made by the engine. Identifiable information sources contribute to accountability. If a decision made by the engine is questioned or challenged, being able to trace back to the sources helps identify any biases, errors, or questionable data that may have influenced the outcome.

Understanding the information associated with data sources allows for ongoing monitoring and updates to correct errors, improve data quality, and enhance the overall performance of the engine. For end-users or non-experts interacting with the engine, having insights into the information sources aids in understanding how the system works. Users are more likely to engage with and trust a system when they can comprehend the basis for its decisions, even if they don't have technical expertise. 

Finally, ethical considerations are increasingly important as AI systems impact various aspects of society, and having visibility into information sources helps ensure fairness and prevent discrimination. Some industries and regions may have regulations that mandate transparency and explainability in AI systems. Clear documentation of information sources aids in compliance with such regulations.

\subsubsection{Dynamic Information}

Dynamic information sources encompass all sources that change in execution time. Thus, there are two types: visual information, which uses high-level viewers to present information; and logs, which are produced by the software components of the robot.

\begin{itemize}
    \item [R1.1] Visual Information: the robotics systems usually have several viewers that allow them to get insights into what is happening. Therefore, these viewers present the robot's behaviors using a high-level representation, which lets developers and human operators understand what is doing the robot. Some examples of these viewers are the visualizers of finite state machines, for instance, YASMIN \cite{yasmin}; and behavior trees, like Groot \cite{githubGitHubBehaviorTreeGroot}. Moreover, there are other visualizers such as the ROS computational graph, which illustrates the ROS nodes and topics; or even more advanced visualized like Foxglove\cite{foxglove}, which is an open-source visualization and debugging tool for your robotics data. 

    \item [R1.2] Logs: logs are a common information source that can be used to explain the robot's behaviors. For instance, in previous works \cite{gonzalez2023using}, we have used logs along with Large Language Models (LLMs) to produce explanations in autonomous robots. However, the more software components are run in the robot, the more logs are produced. To treat this, in our recent work \cite{sobrínhidalgo2024explaining}, we leverage the Retrieval Augmented Generation (RAG) \cite{lewis2020retrieval}, which is a prompt engineering \cite{sahoo2024systematic} technique employed to retrieve only the related logs.

    \item [R1.3] Parameters: Parameters are the configurable settings of software components, capable of being adjusted during execution. For example, the maximum speed utilized in navigation can be modified to suit various environmental conditions, thereby adapting the robot accordingly. Furthermore, in machine learning and deep learning models, their weights can be altered during runtime. An instance of this is adjusting object detection weights dynamically to identify different objects. It is essential to consider all such modifications to elucidate the behaviors exhibited by the robot.
\end{itemize}

\subsubsection{Static Information}

Static information sources encompass all information that does not change during execution time. Thus, there are four types: configuration files, used to configure the software components of the robot; datasets, used to train the machine learning models; machine learning models; and algorithms.

\begin{itemize}
    \item [R2.1] Configuration files aim to configure some software behavior of autonomous robots. These files, while remaining static throughout execution, wield significant influence over the robot's operational dynamics. An illustrative instance lies within the navigation domain, exemplified by the usage of configuration files in ROS 2's Navigation2 framework \cite{macenski2020marathon2}. Within Navigation2, a configuration file is essential for delineating crucial navigation parameters, such as maximum velocity, global planner specifications for path generation, and the selection of controllers for local path planning.
    
    \item [R2.2] Datasets stand as another foundational element, providing static information essential for training machine learning models embedded within autonomous robots. These datasets, immutable during execution, play a vital role in shaping the behaviors and decision-making capabilities of robots, rendering them indispensable in the explanation of robotic actions.
    
    \item [R2.3] Machine learning models, including deep learning architectures, constitute static repositories of knowledge integral to the functioning of autonomous robots. Immutable in their parameterization throughout execution, these models wield substantial influence over a robot's perception and decision-making processes. For instance, the integration of YOLOv8 \cite{Jocher_YOLO_by_Ultralytics_2023,yolov8_ros_2023} enables robots to glean real-world information, highlighting the indispensable role of such models in robotic systems.
    
    \item [R2.4] Algorithms encompass the static software components entrenched within autonomous robots, governing various functionalities without undergoing changes during execution. These algorithms span a spectrum of domains, including navigation systems such as plugins loaded in Nav2 , finite state machines exemplified by YASMIN~\cite{yasmin} , and myriad other essential robotic components. 
\end{itemize}

The static nature of these elements does not avoid having a set of parameters associated with changing their behavior over time (R1.2). This characteristic has to be tracked in order to guarantee transparency on robot behaviors. 

\subsection{RIFT}

The Robotic Interpretability Framework Toolkit (RIFT) is a comprehensive set of tools, methodologies, and components designed to facilitate the development, implementation, and assessment of decision-making processes running in an autonomous robot. 
Interpretability in the context of robotics refers to the degree to which humans, including users, operators, or other stakeholders, can understand and make sense of the decisions and actions made by an autonomous robotic system. It involves providing clear and comprehensible information about the robot's behavior and the systems that generate that behavior. The level of detail of inner models, static/dynamic information and the complexity of the system are directly associated with the Interpretability degree and need to be measured. Thus, it should be investigated in further research in order to facilitate transparency and foster trust between the machine and its human counterparts.


The main problem with interpretability is that it is not an objective concept and is broadly used indistinctly with explainability. Interpretability depends on the user receiving the data. Information can be presented in different forms (e.g., logs, graphs, or natural language). However, the same data can be understood for one user and incomprehensible for another. This variability depends on the user's level of knowledge and familiarity with the data format. In that direction, new research lines based on Large Models can help with this interpretation:
Interpretability is crucial for ensuring trust in the functioning of the robot and its decisions, however, it should be intelligible to those interacting with or impacted by its actions. Thus, it is important that any information provided to the end-user is adapted to its knowledge, as defined in R0.2, and therefore, RIFT is highly linked with REFT.


There are two main requirements associated with RIFT, Model Interpreter and General Purpose Translator, which are associated with REFT and presented in the next subsection.

\begin{itemize}
  
        \item [R3.1] The use of components to explain predictions using images, text, or tabular data can provide insight into the cognitive abilities of the robot. Its aim is to say why and how a robot model or component is generating a prediction and why. 
  
\end{itemize}

For instance, applying these models to the text generated by system logs can identify the most relevant messages during an action.  However, the community should start applying in robotics the same tools that the AI community is doing, tools like SHAP (SHapley Additive exPlanations)~\cite{Lundberg:2017} or LIME (Local Interpretable Model-agnostic Explanations)~\cite{Ribeiro:2016} can reveal which elements within an image are pertinent for making a prediction. It is necessary for these models to offer a metric or explanatory representation that reflects the process used to attain the results.

Besides, the use of the GP Translator would be mandatory to enhance the understandability of the evidence of these components. Thus, it is necessary to follow the three phases revised in Anjomshoae et al.~\cite{anjomshoae2019explainable}: Explanation Generation, Explanation communication and Explanation Reception.

\subsection{REFT}

This toolkit provides tools, resources, or methodologies to facilitate the creation and integration of explanations in robotic decision-making processes. 
Typically, these processes are executed non-transparently to the user. To obtain these explanations, it is necessary to process the data collected by the sources of information independently using tools based on explainability algorithms or Machine Learning. These models can generate metrics, provide visualization of the agent's internal state, or detect relevant events, which may explain the functioning of the agent's cognitive abilities such as perception, deliberative thinking, or action control.
With this approach, the aim is to achieve explainability so that subsequently, an expert agent or an interpretability tool can generate an intelligible outcome for the user.
At this stage, we have defined just two main components, the model interpreter and the general purpose interpreter.

\begin{itemize}

    \item [R4.1] General Purpose Interpreter: it refers to a versatile component or system that is designed to interpret and process a wide array of information sources and types within the robot's information sources and end-user requirements. The primary goal is to enable the robot to understand and respond to diverse forms of data, adapting it to an intelligible format to interact effectively with the end-users and other components within the system.
  
    \item [R4.2]  Visual Interpreter: it refers to a versatile component or system that is designed to interpret and process mainly visual information sources. The primary goal is to enable the robot to provide information about the images and adapt the caption effectively to end-users. This component hones its capabilities to extract meaningful insights from images, enabling the robot to deliver pertinent information and tailor captions effectively for the benefit of end-users. 
\end{itemize}

The General Purpose Interpreter component is necessary in both cases, REFT and RIFT, and it is associated with R0.2, where it is a requirement to present information to end-users. The concept revolves around presenting an explanation in the form of a collection of statements encompassing facts, beliefs, rules, contextual information, clarifications of causes, and potential consequences and should be aligned to the current state-of-the-art designs~\cite{anjomshoae2019explainable}.

\subsection{RAFT}

Defining a "Robotic Accountability Framework Toolkit" (RAFT) involves developing a comprehensive set of guidelines, tools, and resources to ensure ethical, responsible, and transparent design, development and deployment of robotic systems. This toolkit aims to address various aspects of robotic accountability, covering ethical considerations, legal frameworks, safety measures, transparency, cybersecurity and more. Thus, it is a requirement for the design of a component designed to function similarly to an aircraft Flight Data Recorder. This component has to record sensor data sources, actuator behavior and relevant internal status information about the robots. RAFT's idea is to help individuals to perform accident and incident Investigations, and therefore, the role a blackbox plays is a cornerstone in those robots moving outside laboratories and controlled arenas. 

\begin{itemize}
    \item [R5.1] A standardized approach to generate data recording should facilitate consistent and efficient accident and incident investigations involving robots. Besides, it has to present interoperability and adaptability mechanisms. 

\end{itemize}

The market (Companies such as Alias Robotics have already proposed solutions) or the community can although the proposals should not be unique, they should share standard mechanisms of recording and communication with alternative systems\cite{winfield2022ethical}. 
A standard specification, as the one proposed by Winfield~\cite{winfield2022ethical}, enables the sharing and adaptation of black box implementations across different robotic platforms, fostering interoperability and encouraging manufacturers to develop them.


\begin{figure*}[ht]
    \centering
    \vspace{2mm}
    \includegraphics[width=0.99\textwidth]{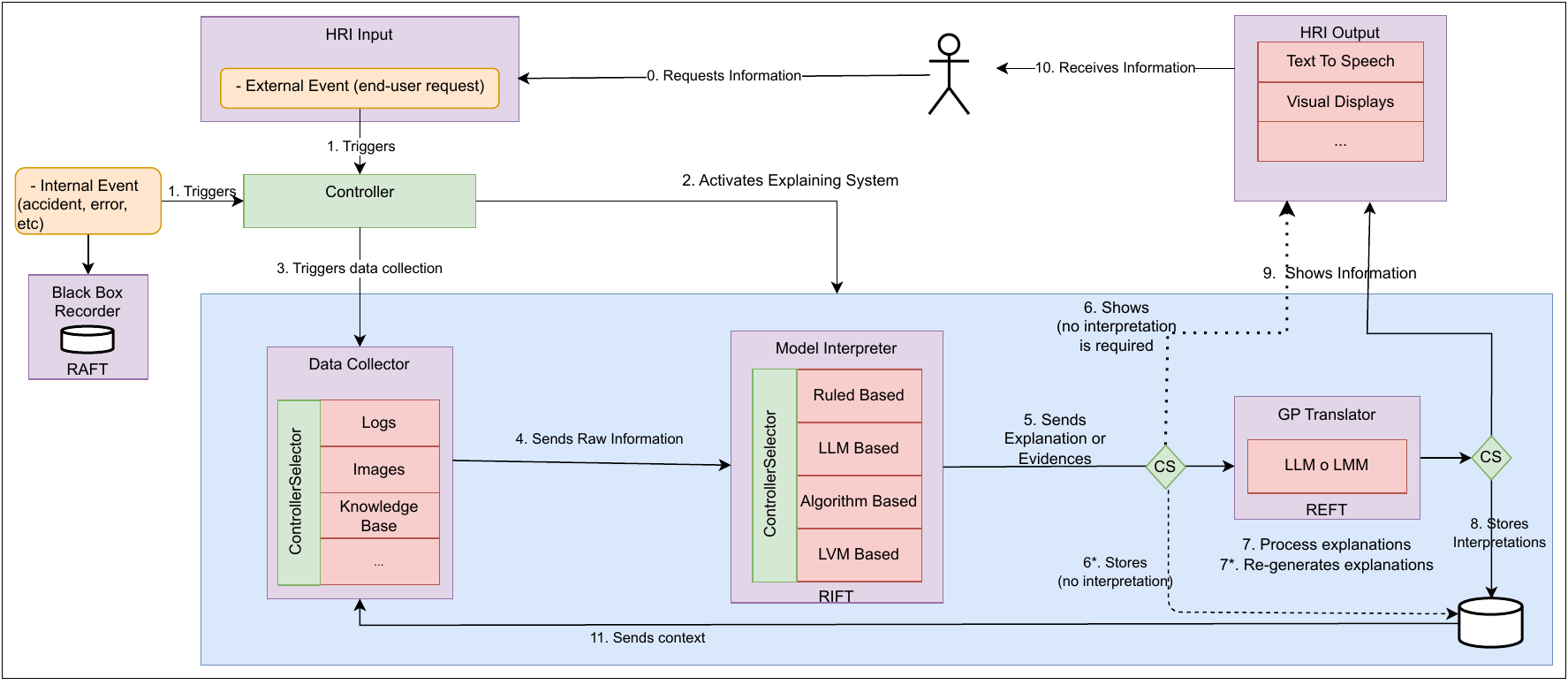}
    \caption{Example of ROXIE concepts integration in a real robot. Here it is presented as the XENIA system. }
    \label{fig:XENIA}
\end{figure*}

\section{PROOF OF CONCEPT}
\label{sec:TOOLKIT}

In the development of an explicable robotic system, the integration of versatile tools plays a pivotal role in ensuring the system's ability to effectively perform explanations. This section delves into the practical application of previous concepts within a robotic system, exploring state-of-the-art solutions employed in our research.

\subsection{Description}

Before defining the explanation flow in our robots, it is mandatory to define the temporal window. Afterward, it would be easy to associate with tools. There are two perspectives for an explanability system online (real-time) and offline (non-real-time) explainable system. Adding a ROXIE to a real-world robot could add new loads that complicate robot performance in public spaces. 

The online approach would be faced from two sides:
\begin{enumerate}
    \item Inherent ROXIE System: Real-time information presentation. This is the classic approach to checking what is happening in the robot. It is usually based on visualization tools such as RVIZ, GROOT or YASMIN viewer. 

    \item Post-hoc ROXIE System: Real-time analysis, decoupling from the audited system, and privacy measures are also desirable features in any accountability approach. Authors in \cite{Fernandez2024} propose an Accountability as a Service (AaaS) framework based on recording and filtering of the system calls and their arguments executed by the monitored system. In addition, this solution includes end-to-end encryption, thus guaranteeing secure logging of all executed events while improving the analysis process.
\end{enumerate}

On the other hand, the offline system uses previous systems based on data stored usually based on rosbags and accountability systems (black box).

\subsection{Flow}

In the domain of autonomous robotics, the seamless interaction between humans and machines hinges upon the ability of robotic systems to provide transparent and comprehensible explanations for their actions. Central to achieving this objective is the development and integration of a ROXIE approach, in this case exemplified by the acronym XENIA. This section elucidates the sequential steps involved in generating and presenting explanations within a robotic system. By dissecting each stage of the process, this section aims to illuminate how XENIA's components synergize to enhance interpretability, foster user trust, and facilitate meaningful human-robot interaction. Figure~ \ref{fig:XENIA} presents our initial approach to deploying in robots.


\begin{enumerate}
    \item \textbf{User requests information:} At the core of the system's functionality lies the user's query, initiating a sequence of events leading to the delivery of explanations. This initial interaction serves as the catalyst for the system's operation, prompting engagement with various components to fulfill the query effectively. However, there are also internal events that would trigger the explainability activation
    
    \item \textbf{System triggers XENIA (eXplicabilidad EN sIstemas Artificiales):} Upon receiving the event query, the system activates the XENIA framework, designed to enhance transparency and interpretability in artificial systems. This initiates subsequent stages of explanation generation and interpretation. Based on lifecycle nodes, the system is activated or deactivated. 
    
    \item \textbf{Activation of explaining system:} With XENIA activated, the Explaining System springs into action, orchestrating the information flow and processes necessary for generating coherent explanations. This step marks the formal start of explanation generation to provide clear insights into decision-making.
    
    \item \textbf{Triggers data collection:} To ensure comprehensive coverage, the system activates dormant data collection processes and gathers non-default sources. By seeking additional data streams, the system aims to enrich its understanding of the query context.
    
    \item \textbf{System sends data to RIFT:} As data collection progresses, the system transmits gathered information to the Robotic Interpretation Framework Toolkit (RIFT). Here there is a Model Interpreter that processes incoming data streams, extracting insights to inform subsequent stages. Figure \ref{fig:LRP_SELFAIR}, the utilization of Layerwise Relevance Propagation (LRP) on a YOLOv8 model trained with dataset X is depicted. This figure illustrates the areas that influence the detection process and to what extent they affect it. This image is presented as is to the end-user for showing information. However, it has a complicated explanation for a generic end-user. 
    
\begin{figure}[ht]
    \centering
    \includegraphics[width=0.48\textwidth]{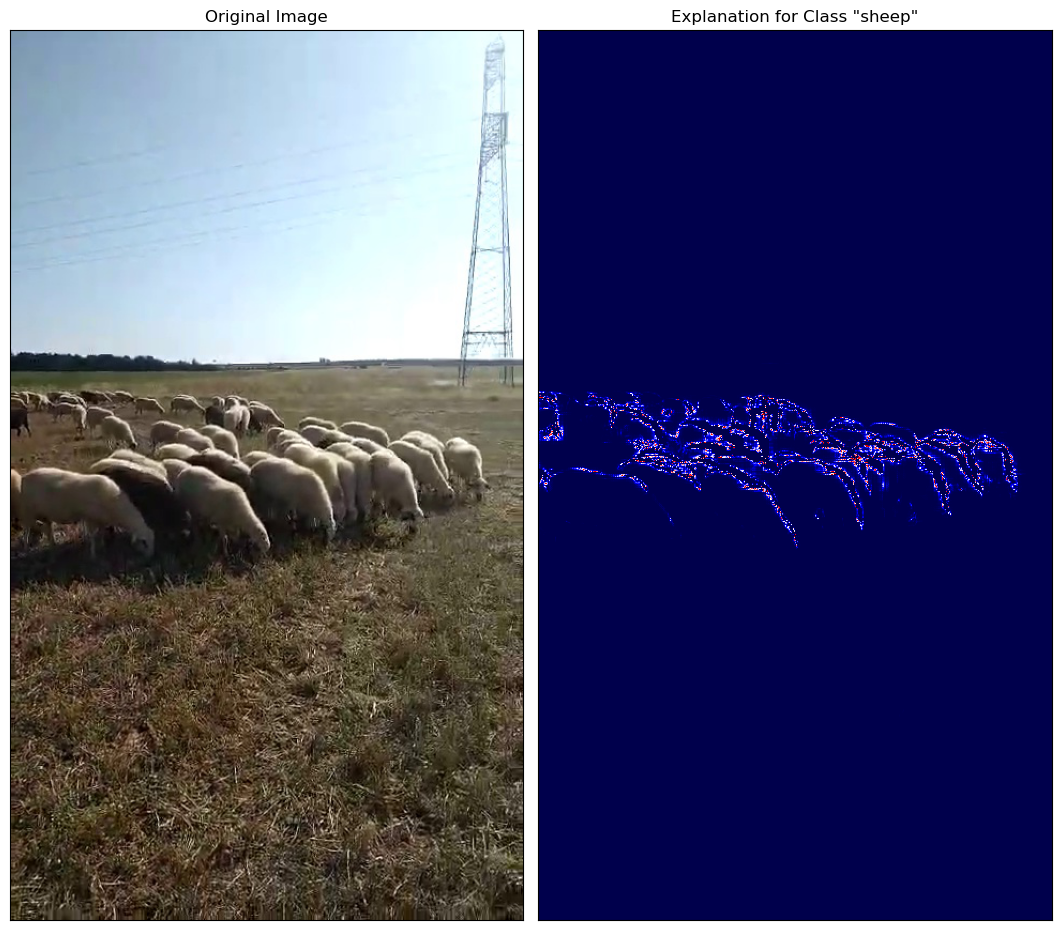}
    \caption{LRP applied to a YOLOv8 model trained on the dataset. If this information is provided as is to non-technical end-users, it will be necessary to measure the level of understanding of the recognition process}
    \label{fig:LRP_SELFAIR}
\end{figure}

    \item \textbf{Controller Selector manages data flow:} Upon receiving processed data from RIFT, the Controller Selector orchestrates distribution within the system. It discerns between data requiring immediate transmission to the Human-Robot Interface (HRI), data for storage, and data for further interpretation by the Robotic Interpretability Framework Toolkit (RIFT). 
    
    \item \textbf{REFT processes information:} REFT analyzes received information, applying algorithms to derive explanations. It distills complex information into actionable insights comprehensible to users. The description and empirical approach can be found in \cite{fernándezbecerra2024enhancing}.
    
    \item \textbf{RIFT information stored in database:} Recognizing the value of preserving insights, the system stores information generated by RIFT in its database. This repository serves as a resource for archiving past interpretations. In our case a General Purpose Translator. 
    
    \item \textbf{Send Information:} Synthesized explanations are transmitted to the Human-Robot Interface (HRI) Output module. This represents the culmination of interpretability efforts, preparing interpretations for users.
    
    \item \textbf{Information presented to user:} From two different perspectives with or without the interpretive process complete, the system presents information to the user. Through visual displays, auditory cues, or textual summaries, the system tailors the presentation for optimal comprehension.
    
    \item \textbf{Previous explanations inform data collection:} The system leverages past explanations stored in its database to inform subsequent data collection iterations. Drawing upon historical context, the system enhances its ability to anticipate user needs and refine interpretive capabilities.
\end{enumerate}

\subsection{Toolkit}

In this study, we aim to incorporate tools into the architecture of our robots to enhance how users receive information and to measure their understanding of it. Table~\ref{tab:tools} showcases many of the tools we are currently utilizing associated to previous flow and the requirements defined above. 

\begin{table}[ht]
\caption{Tools used in different stages of this research used for explaining robot behaviors to end-users. \label{tab:tools}}
\resizebox{.5\textwidth}{!}{
\begin{tabular}{|l|l|l|l|}
\hline
Component & Req.   & Tools                             & Cite                                  \\ \hline
All & R0.1          & ROS SealFS, inmutable BBR                  & \cite{soriano2021sealfs,Ruffin2019}   \\ \hline
All  &  R0.2        & llama\_ros, LLaMA                        & \cite{llama_ros_2023,llama2}          \\ \hline
All  & R1.1         &  ROS 2                            & \cite{macenski2022robot}              \\ \hline
All & R1.2          &  GROOT, YASMIN viewer, Foxglove  & \cite{githubGitHubBehaviorTreeGroot,yasmin, foxglove}            \\ \hline
All & R1.3          &  ROS 2   & \cite{macenski2022robot}            \\ \hline
All &  R2.1         &  yaml, .config, .csv, .json, launcher      &  -          \\ \hline
All &  R2.2         & HuggingFace, YOLOv8       & \cite{Jocher_YOLO_by_Ultralytics_2023,hugging}   \\ \hline
All &  R2.3         & HuggingFace, YOLOv8       & \cite{Jocher_YOLO_by_Ultralytics_2023,hugging} \\\hline
All &  R2.4         & TEB Planner                       & \cite{s21248312}                       \\ \hline
REFT,RIFT  & R3.1   & SHAP, LIME, CAPTUM, LRP, easy\_explain           & \cite{Lundberg:2017, Ribeiro:2016, kokhlikyan2020captum, easyexplain}    \\ \hline
REFT,RIFT & R3.2    & CAPTUM, AD-HOC                    &                                       \\ \hline
REFT & R4.1         & LLaVA, Stable Diffusion           &   \cite{liu2023improved, podell2023sdxl}                               \\ \hline
RAFT &  R5.1        & EBB Winfield, AaaS                &  \cite{winfield2022ethical,Ruffin2019,Fernandez2024}           \\ \hline
\end{tabular}
}
\end{table}

\section{CONCLUSIONS}
\label{sec:CONCLUSIONS}

It becomes evident that the pursuit of interpretability and explainability in robotics and machine learning stands as a paramount endeavor. Thus, the interpretable robot should be able to generate decisions based on humanly understandable rules and the explainable robot refers to the agent that has explicit explanations and justifications for the decisions and actions taken. The REFT aims to communicate the reasoning of any robot technology to end users, while RIFT focuses on increasing transparency so users will understand precisely why and how a model generates its results.

Looking ahead, there are several avenues for further exploration and refinement. Firstly, ongoing efforts to enhance the efficiency and scalability of ROXIE, including systems like RAFT, REFT and RIFT will be crucial for accommodating increasingly complex robotic systems and diverse user needs in real public spaces attending normative rules.  Additionally, exploring novel techniques for integrating interpretable and explainable components into cognitive architectures promises to unlock new possibilities for transparency by design. 

\addtolength{\textheight}{-11cm}   





\section*{ACKNOWLEDGMENT}

This work has been partially funded by an FPU fellowship provided by the Spanish Ministry of Universities (FPU21/01438) and  DMARCE Grant PID2021-126592OB-C21 funded by MCIN/AEI/10.13039/501100011033; and by the Recovery, Transformation, and Resilience Plan, financed by the European Union (Next Generation) thanks to the TESCAC project (Traceability and Explainability in Autonomous Systems for improved Cybersecurity) granted by INCIBE to the University of León. We want also to acknowledge Dagstuhl Seminar 23371 "Roadmap for Responsible Robotics" for their insightful conversations to improve trust in robotics for public spaces.

 \bibliographystyle{IEEEtran}
 \bibliography{IEEEfull}

\end{document}